\documentclass[preprint]{elsarticle}
\usepackage{lineno}
\modulolinenumbers[1]
\usepackage[english]{babel}
\usepackage{graphicx}
\usepackage{longtable}
\usepackage{caption} 
\usepackage{amsmath,amssymb}

\journal{Physica A}
\begin{document}

\begin{frontmatter}

\title{Word frequency-rank relationship in tagged texts}

\author[ac]{Andrés Chacoma}
\ead{achacoma@famaf.unc.edu.ar}
\address[ac]{Instituto de Física Enrique Gaviola, Consejo Nacional de Investigaciones Científicas y Técnicas and Universidad Nacional de Córdoba, Ciudad Universitaria, 5000 Córdoba, Pcia.~de Córdoba, Argentina}

\author[dz]{Damián H.~Zanette\corref{mycorrespondingauthor}}
\cortext[mycorrespondingauthor]{Corresponding author}
\ead{zanette@cab.cnea.gov.ar}
\address[dz]{Centro Atómico Bariloche and Instituto Balseiro, Comisión Nacional de Energía Atómica and Universidad Nacional de Cuyo, Consejo Nacional de Investigaciones Científicas y Técnicas, Av.~Bustillo 9500, 8400 San Carlos de Bariloche, Pcia.~de Río Negro, Argentina}

\begin{abstract}
We analyze the frequency-rank relationship in sub-vocabularies corresponding to three different grammatical classes ({\em nouns}, {\em verbs}, and {\em others}) in a collection of literary works in English, whose words have been automatically tagged according to their grammatical role. Comparing with a null hypothesis which assumes that words belonging to each class are uniformly distributed across the frequency-ranked vocabulary of the whole work, we disclose statistically significant differences between the three classes. This results point to the fact that frequency-rank relationships may reflect linguistic features associated with grammatical function. 
\end{abstract}


\end{frontmatter}

%\linenumbers

\section{Introduction: Frequency-rank relationship and Zipf's law}

The relationship between word frequencies and ranks in written texts is, beyond any doubt, the most thoroughly studied among the statistical properties of language usage \cite{Pianta}. Following G.~K.~Zipf's prescription \cite{Zipf}, the {\em rank} of a word is defined as its position in a list where all the different words of a text are arranged in decreasing order by their number of occurrences, or {\em frequency}. Zipf himself pointed out the ubiquitous regularity according to which the frequency of a word is approximately proportional to the inverse of the rank, at least for large ranks and low frequencies --a systematic feature which came to be known as {\em Zipf's law}. The linguistic relevance of this regularity has been discussed in connection with the reinforcement in the use of certain words in detriment of others, as a text progresses and its semantic contents grow \cite{Simon,RZ}. The underlying mechanism is not unlike preferential attachment \cite{PA} or sample space reducing \cite{SSR}, which are well known to generate algebraic (power-law) functional relations and distributions. 

Thanks to novel advancements in techniques of machine learning and natural language processing \cite{nltkb}, it has lately become possible to automatically annotate digitized texts, assigning a tag to each word token, which indicates its grammatical role in the corresponding sentence. This possibility opens a plethora of new lines of computational research on the statistical properties of language, now discerning between word classes with different lexical functions. In a recent contribution \cite{nos}, we have presented a statistical analysis of vocabulary growth --namely, appearance of new words as a text gets longer-- in a corpus of  automatically tagged literary works. Scaling properties in the relationship between vocabulary size and text length are encompassed by Heaps' law \cite{Heaps,H2}. Encouragingly, we have found that these properties are different for each considered grammatical class, which suggests that meaningful linguistic information is enclosed in such features. 

Here, we extend this kind of analysis to the frequency-rank relationship in the same collection of tagged texts. It should be understood, however, that it is not our aim to assess whether Zipf's law holds within grammatical classes. Instead, we are interested in whether and how the lexical function of each class has a statistically discernible effect on the frequency-rank relationship. In the next section, we present the corpus of texts on which we work, and the null hypothesis that we use for comparison to determine the statistical significance of our results. Quantitative measures to evaluate the difference between real frequency-rank relationships and our null hypothesis are proposed and computed in Sect.~\ref{S3}, where we show that significant differences appear in the relationships corresponding to each grammatical class. In Sect.~\ref{S4}, we sketch an interpretation of the results, by means of a simple model which highlights the disparate distribution of words belonging to each grammatical class within the whole vocabulary of each text. Finally, our main results are summarized in Sect.~\ref{S5}.

\section{Materials and methods: Tagged corpus and null hypothesis}

\subsection{The corpus}

In this paper, we study a corpus of $75$ literary works in English, authored by six well-known British and North American writers of the nineteenth and twentieth centuries. The individual text lengths vary between $N\approx 800$ to $3.5 \times 10^5$ word tokens, with vocabularies between $V\approx 300$ to $2.2 \times 10^4$ word types in size. Digital versions of the texts were downloaded from Project Gutenberg \cite{guten} and Faded Page \cite{faded}. The list of works in the corpus,  indicating author, title, publication year, length, and vocabulary size, can be found in a recent open-access publication \cite{nos}. The collection of texts used in our study is available online in a public repository \cite{dryad}.

As a first step, content not belonging to the original works was removed manually from each file. Then, each text was automatically processed using the Natural Language Toolkit library (NLTK)  \cite{nltk,nltkb}. The crucial stage in the process is tagging, which consists of a classification of all the word tokens into the $35$ lexical categories recognized by the NLTK part-of-speech tagger. NLTK tagging uses a combination of techniques, driven by hidden Markov model training \cite{HMM}. In order to increase the statistical significance of sampling, we have aggregated the $35$ categories into just three broad grammatical classes. These are {\em nouns}, which include singular and plural common and proper nouns, and personal pronouns; {\em verbs}  in all persons and tenses; and {\em others}, comprising the remaining categories. The result of processing each text is a sequence of word tokens, each of them with a new attribute indicating its grammatical class. For clarity in the presentation, we call each one of the three classes a {\em tag}. 

\subsection{Frequency-rank relationship within tags} \label{FRRT}

Given the tagged vocabulary of each text, our focus is here put on the frequency-rank relationship within the sub-vocabulary corresponding to each one of the three tags. Our working hypothesis is that this relationship may shade light, from a statistical viewpoint, on the different linguistic role of the three grammatical classes. 

Let $w_r$ be the word type of rank $r$ in the whole vocabulary (i.e., the $r$-th most frequent word) and $f_r$ its number of occurrences (i.e., its frequency) in the text. Consider now any subset of the vocabulary --namely, any sub-vocabulary-- that contains $w_r$. If the word types in this sub-vocabulary are ranked by their frequency, the rank $r'$ of $w_r$ will necessarily satisfy $r'\le r$. Moreover, given two word types  $w_{r_1}$ and $w_{r_2}$ with ranks $r_1 < r_2$ in the original vocabulary, their ranks in the sub-vocabulary (provided that both of them belong to it) will satisfy $r'_1 < r'_2$. These remarks show that the frequency-rank relation in the sub-vocabulary is shifted to lower ranks with respect to that of the whole vocabulary, and that the original relative ordering of word types is preserved.

\begin{figure}[h]
    \centering
    \includegraphics[width=0.8\textwidth]{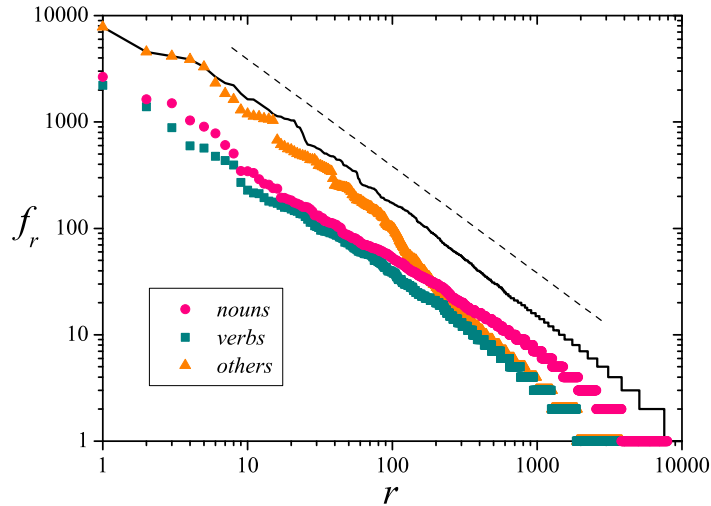}
    \caption{Frequency-rank relationship (number of occurrences $f_r$ vs.~rank $r$) in the tagged version of A. Huxley's {\em Eyeless in Gaza}. The full curve stands for the whole vocabulary, while symbols correspond to the sub-vocabularies formed by each tag ({\em nouns}, {\em verbs}, and {\em others}). The dashed straight line has slope $-1$. }
    \label{fig1}
\end{figure}

The full curve in Fig.~\ref{fig1} shows the frequency-rank relationship for the whole vocabulary of one of the tagged texts in the corpus, A.~Huxley's {\em Eyeless in Gaza}, for which $V=19068$. Overall, the relationship is the same as found for non-tagged texts, closely following Zipf's law, $f_r \propto r^{-1}$, for intermediate ranks. In fact, a linear fit for $10<r<1000$ over the log-log plot returns a slope equal to $-1.064 \pm 0.001$ with a correlation coefficient above $0.999$. 

In the same figure, frequency-rank relationships for the sub-vocabularies corresponding to each tag are shown by symbols. The sub-vocabulary sizes are $V_{nouns}=9401$, $V_{verbs}=4715$, and $V_{others}=4952$. Note that, for the first few values of $r$, data for {\em others} and the whole vocabulary coincide. This is due to the fact that, as expected, the most frequent words in the vocabulary ({\em the}, {\em of}, {\em and}, {\em to}, {\em a}) are function words --i.e. words without lexical meaning \cite{fw}-- which are tagged as {\em others}. The most frequent word in  {\em nouns} is {\em he}, with rank $r=6$ in the whole vocabulary, followed by {\em it} ($r=10$), {\em she} ($r=12$), {\em you} ($r=21$), and {\em I} ($r=22$). As for {\em verbs}, the most frequent words are {\em was} ($r=8$),  {\em had} ($r=13$), {\em be} ($r=23$), {\em were} ($r=28$), and {\em said} ($r=30$). Consequently, for low ranks, data for {\em nouns} and {\em verbs} lie below those corresponding to {\em others}.  We point out by passing  that, except perhaps for the tag {\em nouns}, these frequency-rank relationships do not admit a satisfactory power-law fitting over more than a decade in the rank, which raises a doubt  whether Zipf's law holds for the respective sub-vocabularies. 

To disclose whether the frequency-rank relationship within each tag contains information of linguistic relevance, we propose to compare results obtained for the respective sub-vocabularies in each text of the corpus with those obtained for random sub-vocabularies of the same sizes. Our null hypothesis is that they cannot be statistically discerned from each other. Any significant departure, on the other hand, can possibly be interpreted in terms of the different linguistic role of each tag. In the following, we provide exact results concerning the frequency-rank relationship in random sub-vocabularies.   

\subsection{Null hypothesis} \label{2.3}

Consider a text with a vocabulary formed by $V$ different words, where --as above-- the number of occurrences of $w_r$, the word with rank $r$, is $f_r$. From this vocabulary, take a random subset of $V'$ words. What is the probability that $w_r$ shifts to rank $r'$ in the sub-vocabulary? This question can be recast as a handsome problem in probability theory: {\em From an ordered sequence of $V$ distinct elements, $V'$ of them are extracted at random, and their order is maintained. What is the probability $p(r\to r')$ that the element at place $r$ in the original sequence ends at  $r'$ in the resulting sub-sequence?}  Suitable combinatorial arguments make it possible to find that 
\begin{equation} \label{1}
    p(r\to r') =  \left( \begin{array}{c} V \\ V' \end{array}\right)^{-1}
\left( \begin{array}{c} r-1  \\ r'-1 \end{array}\right)
\left( \begin{array}{c} V-r  \\ V'-r' \end{array}\right) .
\end{equation}
As expected, this results is well defined for $V'\le V$ and $r'\le r$ only. Moreover, it is required that $V'-r' \le V-r$, because the number of words with ranks above $r'$ in the sub-vocabulary must be at most equal to that above $r$ in the original vocabulary. For any other combination of $V$, $V'$, $r$ and $r'$, we have $p(r\to r')=0$. Note also that, if $V'<V$,  $p(r\to r')$ is {\em not} normalized with respect to $r'$:
\begin{equation}
    \sum_{r'=1}^r p(r\to r') = \frac{V'}{V} < 1.
\end{equation}
This fact takes into account the possibility that $w_r$ is not chosen to belong to the sub-vocabulary, in which case its contribution to $p(r\to r')$ becomes ``lost''. On the other hand,
\begin{equation}
    \sum_{r=r'}^V p(r\to r') =  1,
\end{equation}
because the word which ends at rank $r'$ in the sub-vocabulary must necessarily come from a rank $r$ between $r'$ and $V$, both inclusive.

Using $p(r\to r')$, it is possible to compute the average value of $r'$ as a function of $r$ --namely, the expected rank of $w_r$  in the sub-vocabulary-- which turns out to be 
\begin{equation} \label{avr}
\langle r'\rangle =\frac{\sum_{r'=1}^r  r' p(r\to r') }{\sum_{r'=1}^r    p(r\to r') } = 1+(r-1)\frac{V'-1}{V-1}. 
\end{equation}
This average is a linear function of $r$, starting at $\langle r'\rangle=1$ for $r=1$ and ending at $\langle r'\rangle=V'$ for $r=V$. Note that $\langle r'\rangle$ coincides with the value of $r'$ expected for $w_r$ provided that the sub-vocabulary is uniformly distributed along the whole vocabulary (see also Sect.~\ref{S4}). The standard deviation of $r'$ can also be exactly calculated:  
\begin{equation} \label{sdr}
\sigma_{r'} =\sqrt{ \frac{(V-V')(V'-1)(V-r) (r-1)}{(V-2) (V-1)^2}}.
\end{equation}
It vanishes at the two ends, $r=1$ and $r=V$, and reaches a maximum at $r=(V+1)/2$.

The expected number of occurrences of the word of rank $r'$ in the random sub-vocabulary is given by a sum of suitably weighted contributions coming from all the words with $r\ge r'$ in the original vocabulary, namely, 
\begin{equation} \label{4}
    \langle f_{r'} \rangle =  \sum_{r=r'}^V f_r  p(r\to r') ,
\end{equation}
where we have taken into account that, as stated above, $f_{r'}= f_r$.
The corresponding standard deviation is $\sigma_{f'} =\sqrt{\langle f_{r'}^2 \rangle-\langle f_{r'} \rangle^2 }$, with
\begin{equation} \label{5}
    \langle f_{r'}^2 \rangle =  \sum_{r=r'}^V f_r^2  p(r\to r').
\end{equation}
Note that, in contrast with the expected rank and its standard deviation, Eqs.~(\ref{avr}) and (\ref{sdr}), $\langle f_{r'} \rangle$ and $\langle f_{r'}^2 \rangle$ depend on the specific collection of word frequencies of each text. It is interesting to mention that, if $f_r \propto r^{-1}$ --namely, if word frequencies strictly follow Zipf's law-- Eq.~(\ref{4}) implies that $\langle f_{r'} \rangle \propto (r')^{-1}$ for sufficiently large values of $V$, $V'$, and $r'$. Under these conditions, therefore, the random sub-vocabulary preserves Zipf's law.

Actually, in our comparison of real tagged texts with the null hypothesis, we do not analyze word frequencies directly, but rather their logarithm $g_r = \log f_r$. In this way --much as when Zipf frequency-rank relationships are plotted in logarithmic scale-- we partially balance the largely disparate contributions of low and high ranks over the interval of variation of frequencies. For the quantities $g_r$, in full analogy with Eqs.~(\ref{4}) and (\ref{5}), we define the mean values
\begin{equation} \label{promg}
    \langle g_{r'} \rangle =  \sum_{r=r'}^V g_r  p(r\to r') , \ \ \ \ \  \langle g_{r'}^2 \rangle =  \sum_{r=r'}^V g_r^2  p(r\to r'), 
\end{equation}
and use them to compute the standard deviation $\sigma_{g'} =\sqrt{\langle g_{r'}^2 \rangle-\langle g_{r'} \rangle^2 }$. For the sake of brevity, in the following we call $g_r$ the {\em log-frequency} of the word type of rank $r$.

\section{Results: Rank and frequency anomalies within tags} \label{S3}

\begin{figure}[h]
    \centering
    \includegraphics[width=0.8\textwidth]{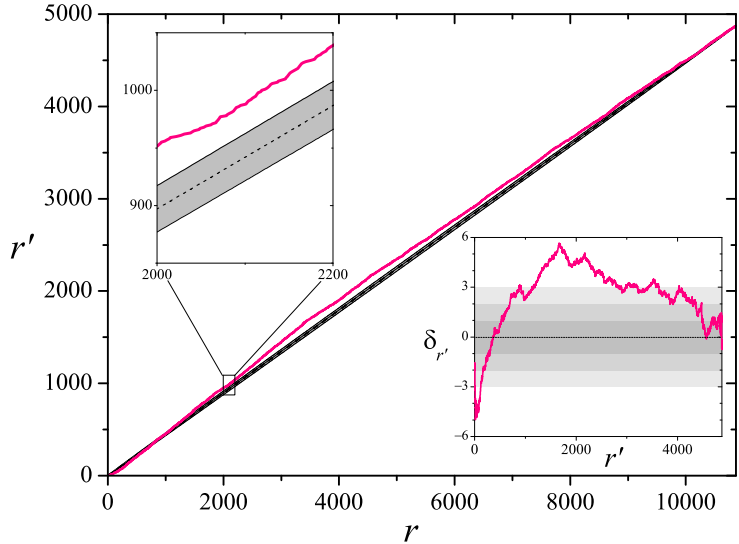}
    \caption{Main panel: Relationship between the ranks $r'$ in the {\em nouns} sub-vocabulary and $r$ in the whole vocabulary for M.~Twain's {\em The Prince and the Pauper}. The colored line corresponds to data for the actual work, and the dashed straight line is the null-hypothesis prediction for the average $\langle r'\rangle$, Eq.~(\ref{avr}). The width of the narrow band around the dashed straight line equals the predicted standard deviation. Upper-left inset: Close-up of the main panel for $2000<r<2200$. Lower-right inset: The rank anomaly $\delta_{r'}$, as given by the first of Eqs.~(\ref{anom}), computed for the same data.}
    \label{fig2}
\end{figure}

\begin{figure}[h]
    \centering
    \includegraphics[width=0.8\textwidth]{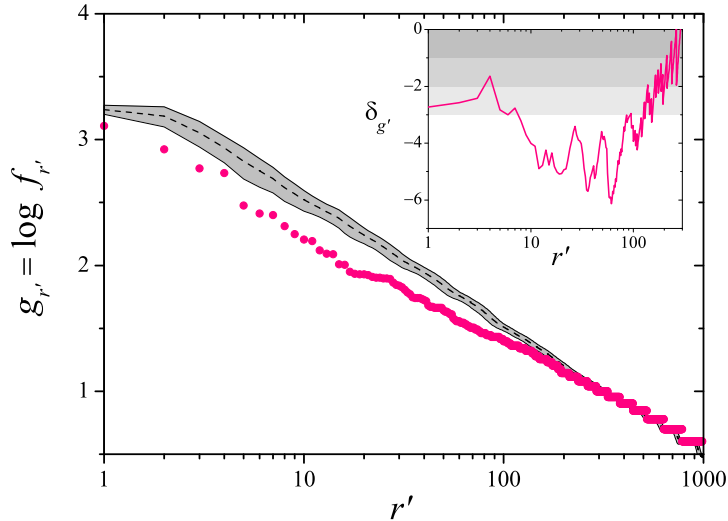}
    \caption{Main panel: The log-frequency $g_{r'}$ (defined as a decimal logarithm) versus the rank $r'$ in the {\em nouns} sub-vocabulary of M.~Twain's {\em The Prince and the Pauper}. Symbols correspond to data for the actual work, and the dashed  line is the null-hypothesis prediction for the average $\langle g_{r'}\rangle$, Eq.~(\ref{avr}). The width of the narrow band around the dashed line equals the predicted standard deviation. Inset: The log-frequency anomaly $\delta_{g'}$, as given by the second of Eqs.~(\ref{anom}), computed for the same data.}
    \label{fig3}
\end{figure}

As an illustration of the comparison between frequencies and ranks in the sub-vocabulary corresponding to a specific tag in a text and its random counterpart, in Figs.~\ref{fig2} and \ref{fig3} we present results for {\em nouns} in M.~Twain's {\em The Prince and the Pauper}. For this work, the whole vocabulary consists of $V=10869$ word types, while the {\em nouns} sub-vocabulary has $V_{nouns}=4873$.

The colored line in the main panel of Fig.~\ref{fig2} shows the rank $r'$ in the {\em nouns} sub-vocabulary versus the rank $r$ in the whole vocabulary. The dashed line stands for the linear functional dependence expected from our null hypothesis, Eq.~(\ref{avr}). As clarified by the close-up shown in the upper-left inset, the dashed line is surrounded by a shaded band, whose width is given by the standard deviation  expected for a random sub-vocabulary of the same size, Eq.~(\ref{sdr}). We see in this example that, although the relationship between $r'$ and $r$ in the actual text roughly follows a linear profile, deviations from the null-hypothesis prediction are statistically significant. In the inset close-up, for instance, the difference between data for the real text and the null-hypothesis average is around twice the standard deviation.     

The main panel of Fig.~\ref{fig3}, in turn, shows analogous results for the log-frequency $g_{r'}$ versus $r'$. In this case, data for the actual text lie systematically below the null-hypothesis prediction and, except in the large-rank tail, well outside the standard-deviation band.

\subsection{Rank and log-frequency relative anomalies}

To quantify the difference between the frequency-rank relationships in the sub-vocabularies corresponding to each tag and their random counterparts, we introduce the {\em relative anomalies} for ranks and log-frequencies,
\begin{equation} \label{anom}
    \delta_{r'} = \frac{r'- \langle r'\rangle}{\sigma_{r'}}, \ \ \ \ \  \ \ \ \delta_{g'} = \frac{g_{r'} - \langle g_{r'} \rangle }{\sigma_{g'}},
\end{equation}
where averages and standard deviations are calculated for the null hypothesis,  as in Sect.~\ref{2.3}. The quantities $r'$ and $g_{r'}$, on the other hand, are those corresponding to the sub-vocabularies of each actual text. Note that both $\delta_{r'}$ and $\delta_{g'}$ are to be computed for each individual tag, and that they are functions of the rank $r'$. We point out, moreover, that $\delta_{g'}$ is independent of the logarithm base used to define $g_r$.

The lower-right inset of Fig.~\ref{fig2} shows the relative anomaly $\delta_{r'}$ versus $r'$ for {\em The Prince and the Pauper}. Horizontal shaded bands bounded by integer values of $\delta_{r'}$ help to appraise the difference between real data and the null hypothesis, relative to the null-hypothesis standard deviation. This difference varies between negative and positive values, up to five or six times as large as $\sigma_{r'}$. Except for the smallest ranks, $r' \lesssim 400$, $\delta_{r'}$ remains mostly positive. For these nouns, therefore, the actual rank in the sub-vocabulary is larger than expected by chance.

Analogously, the inset of Fig.~\ref{fig2} shows $\delta_{g'}$ versus $r'$. In this case, the log-frequency relative anomaly is always negative, showing that the actual values of $g_{r'}$ are below those expected by chance, with a difference up to six times the standard deviation. Note that, in this plot, the horizontal axis is limited to $r'\lesssim 300$. For larger ranks, in fact, $\delta_{g'}$ displays sharp saw-like oscillations around or close to zero. These oscillations originate in the difference between the precise position of the ``steps'' in the tail of the frequency-rank relationship (see Fig.~\ref{fig1}) for real data and the null-hypothesis. Since this artifact  affects the words with the lowest frequencies only, in the following we limit the consideration of  the average anomaly to words with more than ten occurrences, $f_{r'}>10$.

\subsection{Mean anomalies}

In order to condense the information provided by the relative anomalies into a single quantity for each text and tag, we define the respective {\em mean anomalies} as
\begin{equation} \label{avanom}
    \Delta_{r'} = \frac{1}{V'} \sum_{r'=1}^{V'} \delta_{r'}, \ \ \ \ \ \ \ \ 
    \Delta_{g'} = \frac{1}{r'_0} \sum_{r'=1}^{r'_0} \delta_{g'}.
\end{equation}
The sum in $\Delta_{g'}$ is limited to $r'_0$, the maximum rank in the sub-vocabulary for which $f_{r'}>10$. Meanwhile, the respective standard  deviations are
\begin{equation}  
    \Sigma_{r'} = \sqrt{\frac{1}{V'} \sum_{r'=1}^{V'}\left( \delta_{r'} - \Delta_{r'} \right)^2}, \ \ \ \ \ \ \ \ 
    \Sigma_{g'}  = \sqrt{\frac{1}{r'_0} \sum_{r'=1}^{r'_0} \left( \delta_{g'} -\Delta_{g'} \right)^2}.
\end{equation}

\begin{figure}[h]
    \centering
    \includegraphics[width=0.55\textwidth]{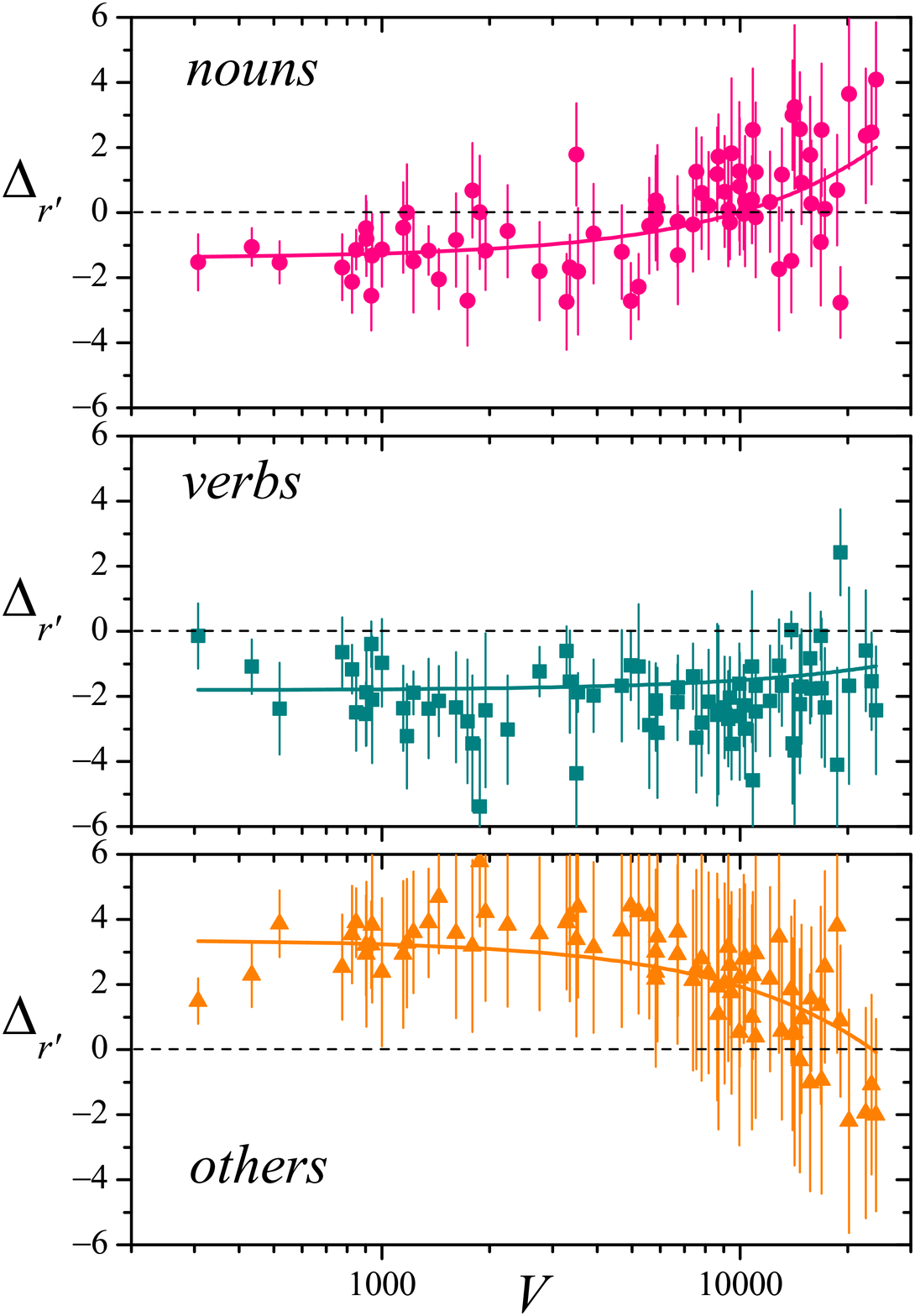}
    \caption{The rank mean anomaly $\Delta_{r'}$ for each text and tag versus the size of the whole vocabulary $V$. Vertical segments indicate the corresponding standard deviation $\Sigma_{r'}$. Curves stand for least-square linear fittings of $\Delta_{r'}$ vs.~$V$.}
    \label{fig4}
\end{figure}

\begin{figure}[h]
    \centering
    \includegraphics[width=0.55\textwidth]{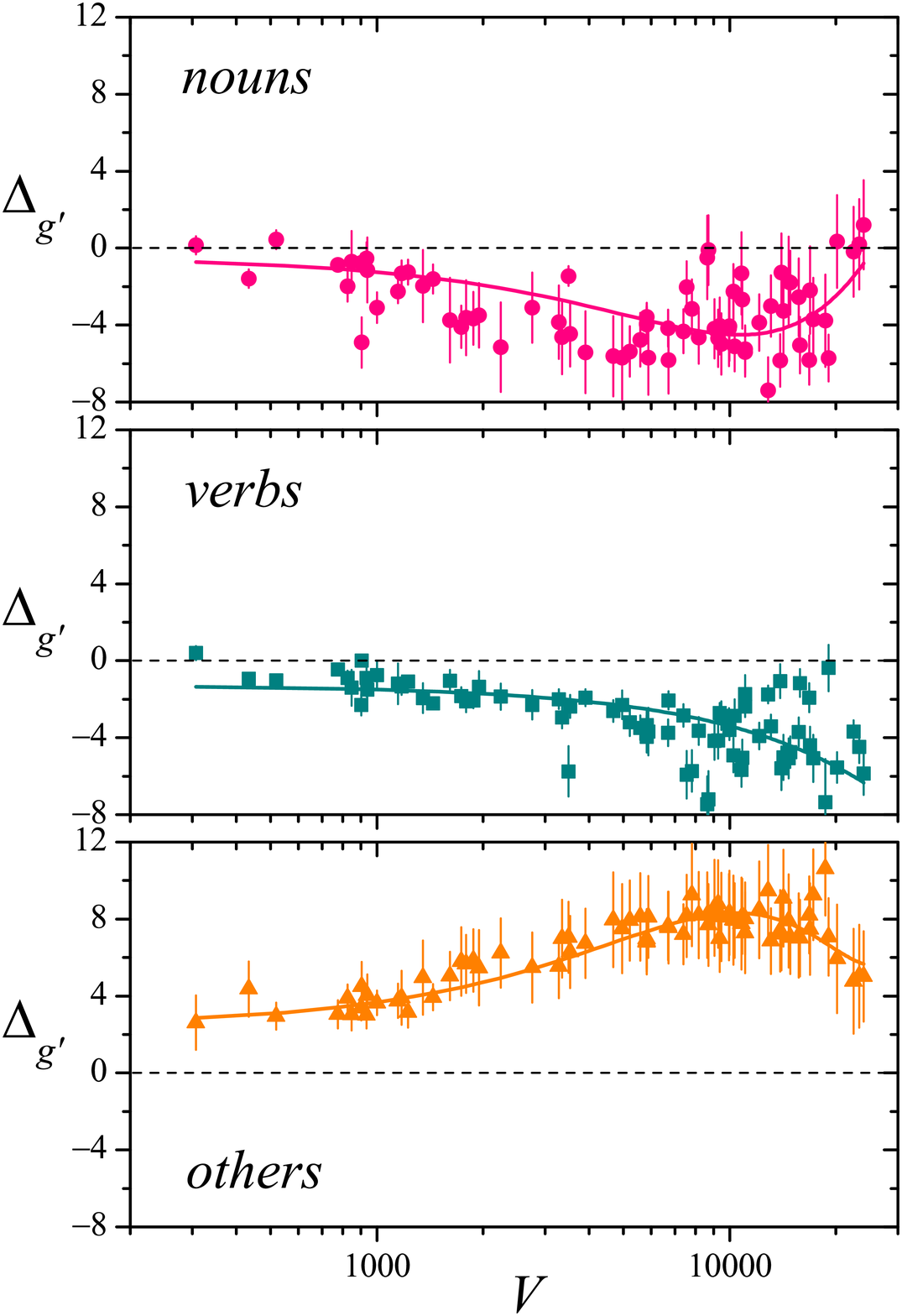}
    \caption{The log-frequency mean anomaly $\Delta_{g'}$ for each text and tag versus the size of the whole vocabulary $V$. Vertical segments indicate the corresponding standard deviation $\Sigma_{g'}$. Curves are polynomial fittings of $\Delta_{g'}$ vs.~$V$, as detailed in the main text.}
    \label{fig5}
\end{figure}

Figures \ref{fig4} and \ref{fig5} show the main results of our analysis, namely, the mean anomalies $\Delta_{r'}$ and $\Delta_{g'}$ for each text and tag, versus the size $V$ of the whole vocabulary of the text. Vertical segments on each dot stand for the deviations $\Sigma_{r'}$ and $\Sigma_{g'}$.

We see in Fig.~\ref{fig4} that the rank mean anomaly $\Delta_{r'}$ reveals a well-defined difference between the statistical behavior of {\em nouns} and {\em verbs} on the one hand, and {\em others} on the other. Specifically, for the smaller vocabularies, $\Delta_{r'}$ is predominantly negative for the former and positive for the latter. Between {\em nouns} and {\em verbs}, in turn, $\Delta_{r'}$  tends to be closer to positive values for the former. As $V$ grows, the mean anomalies for {\em others} and {\em nouns}  approach each other, while for {\em verbs} it remains negative for most texts. As a guide to the eye, we have plotted as lines the least-square linear fittings of   $\Delta_{r'}$  versus $V$ for each tag (which are seen as curves because of the logarithmic scale in the horizontal axis).

Figure \ref{fig5} shows that the difference of {\em others} with the other two tags is more conspicuous when it comes to the log-frequency mean anomaly. In fact,  $\Delta_{g'}$ is always positive for {\em others}. Moreover, it now displays a non-monotonic dependence on $V$, with larger values for intermediate vocabulary sizes. The curve in the plot represents a third-degree polynomial fitting of   $\Delta_{g'}$  versus $V$. For {\em nouns} and {\em verbs}, on the other hand, $\Delta_{g'}$ is almost always negative, and there is no obvious separation between the two tags. Closer inspection, however, discloses a different trend with $V$, with {\em nouns} mirroring the non-monotonic dependence of {\em others}, and {\em verbs} with an overall growth of the absolute value of $\Delta_{g'}$  with the vocabulary size. In the plot, the curves for {\em nouns} and {\em verbs} are  third- and first-order polynomial fittings, respectively. 

\begin{figure}[h]
    \centering
    \includegraphics[width=.8\textwidth]{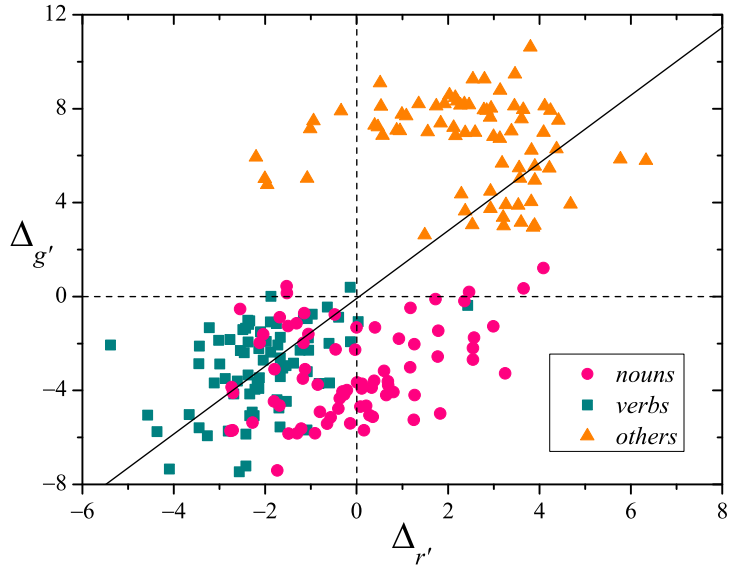}
    \caption{The log-frequency mean anomaly $\Delta_{g'}$ versus the rank mean anomaly $\Delta_{r'}$ for each tag in the $75$ works of the corpus. The straight line stands for the least-square fitting of the whole set of points.}
    \label{fig6}
\end{figure}

Finally,  in Fig.~\ref{fig6} we disregard the information on the the vocabulary size and plot $\Delta_{g'}$ versus $\Delta_{r'}$. Although {\em nouns} and {\em verbs} exhibit a considerable overlapping in the quadrant of negative mean anomalies, the former also occupies a  sizable zone with $\Delta_{g'} <0< \Delta_{r'}$  --as illustrated, in the case of {\em The Prince and the Pauper}, by the insets of Figs.~\ref{fig2} and \ref{fig3}). {\em Others}, in turn, is well separated from the other two tags, as expected. Note the clear positive correlation between the two mean anomalies. The oblique straight line corresponds to the least-square linear fitting of the whole set of points, with a Pearson's correlation coefficient $\rho \approx 0.69$.  

The disparate behavior of the mean anomalies for the three tags is an indication that the present analysis may be  revealing significant statistical features with a linguistic meaning, since the sub-vocabularies of the different lexical categories encompassed by the tags display discernible trends when compared with our null hypothesis. In the next section, we attempt an interpretation of these results by explicitly calculating the mean anomalies for a simple class of sub-vocabulary distributions.

\section{Interpretation: A model sub-vocabulary} \label{S4}

Our conjecture is that, as demonstrated by Fig.~\ref{fig1} in the case of {\em Eyeless in Gaza} (Sect.~\ref{FRRT}), a disparate behavior of the frequency-rank relationship for each tag is related to mutual differences in the distribution of the corresponding sub-vocabularies over the ranked list of words of the whole text. To test this idea, we consider a stylized model consisting of a sub-vocabulary formed by $V'$ words, whose most and least frequent words respectively have ranks $r_{\min}$  and $r_{\max}$ in the whole vocabulary ($r_{\min}<r_{\max}$). We assume moreover that the $V'$ words are uniformly distributed between $r_{\min}$  and $r_{\max}$, so that the word of rank $r'$ in the sub-vocabulary has rank 
\begin{equation} \label{rmod}
    r = r_{\min} + \frac{r'-1}{V'-1} \left(r_{\max}- r_{\min}\right)
\end{equation}
in the whole vocabulary. Equivalently,
\begin{equation} \label{rpmod}
    r' = 1 + \left(r-r_{\min}\right) \frac{V'-1}{r_{\max}-r_{\min}}.
\end{equation}
Note that, if $r_{\min}=1$ and $r_{\max}=V$, this expression for $r'$ coincides with $\langle r'\rangle$ as given by Eq.~(\ref{avr}). 

To estimate the rank mean anomaly $\Delta_{r'}$ for this model sub-vocabulary, we first assume that, in the first of Eqs.~(\ref{anom}), the standard deviation $\sigma_{r'}$ can be replaced by an effective value  $\sigma_{r'}^{\rm eff}$, independent of $r'$. Within this simplifying assumption, and taking into account Eqs.~(\ref{anom}) and (\ref{avanom}), we have
\begin{equation} \label{Drmod}
    \Delta_{r'} = \frac{1}{V' \sigma_{r'}^{\rm eff}} \sum_{r'=1}^{V'} \left(r'-\langle r'\rangle \right) .
\end{equation}
In the sum, the functional relation between the expected rank $\langle r'\rangle$ and the actual rank $r'$ is determined by Eq.~(\ref{avr}), with $r$ given as a function of $r'$ by Eq.~(\ref{rmod}). Our estimation for $\Delta_{r'}$ can be written explicitly, either by exactly calculating the sum in Eq.~(\ref{Drmod}) or by approximating it by an integral. Assuming $V,V'\gg 1$ the two results coincide, and read 
\begin{equation} \label{Drmod1}
    \Delta_{r'} = \frac{V'}{2 \sigma_{r'}^{\rm eff}} \left(1-\frac{r_{\min}+r_{\max}}{V} \right) .
\end{equation}
This result makes apparent that, within the model hypotheses, the sign of $\Delta_{r'}$ is straightforwardly determined by how $V$ compares with $r_{\min}+r_{\max}$. In particular, if both $r_{\min}$ and $r_{\max}$ are sufficiently large, the rank mean anomaly is positive, and {\em vice versa}.

\begin{figure}[h]
    \centering
    \includegraphics[width=\textwidth]{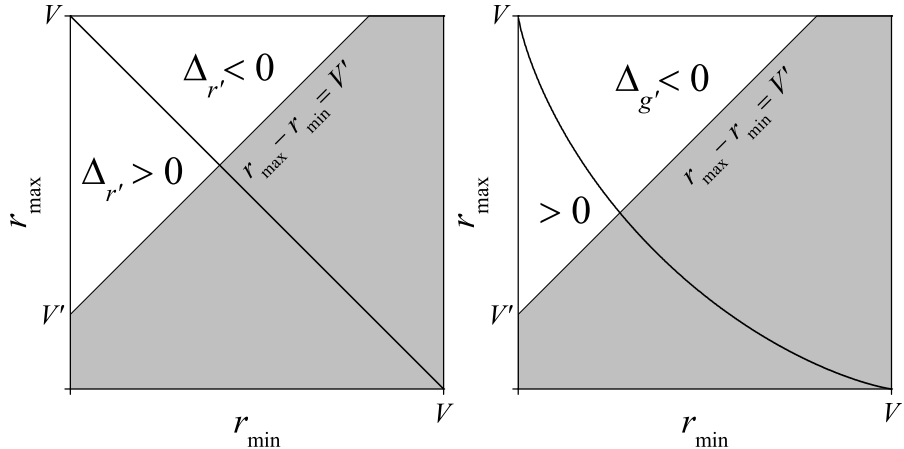}
    \caption{Left panel: Regions in the plane spanned by $r_{\min}$ and $r_{\max}$ where the estimate for $\Delta_{r'}$ given by Eq.~(\ref{Drmod}) is either negative or positive, as indicated by labels. The shaded region below the line  $r_{\max}-r_{\min} =V'$ is forbidden. Right panel: As in the left panel, for $\Delta_{g'}$, Eq.~(\ref{Dgest}). }
    \label{fig7}
\end{figure}

The leftmost panel of Fig.~\ref{fig7} helps assessing this connection more quantitatively. In the plane spanned by $r_{\min}$ and $r_{\max}$, their values are limited to the region with $0<r_{\min}<r_{\max}<V$ and $r_{\max}-r_{\min}>V'$, indicated by the upper-left non-shaded triangle. Within this region, $\Delta_{r'}$ is positive or negative depending on $r_{\min}+r_{\max}$ being less or larger than $V$, as indicated by the labels. On the average, therefore, relatively small or large values of $r_{\min}$  and $r_{\max}$  respectively determine positive or negative values of $\Delta_{r'}$.

In terms of the frequency-rank distribution of words, a negative value for the rank average anomaly --as obtained in our corpus for {\em nouns} and {\em verbs}-- can be interpreted to indicate a sub-vocabulary which is relatively shifted towards larger ranks, namely, a set of words of relatively low frequencies. On the other hand, $\Delta_{r'}>0$ --as is mostly the case for the tag {\em others} in the corpus-- should correspond to a sub-vocabulary with rather high frequencies. This interpretation provides statistical significance to our remark in connection with Fig.~\ref{fig1} (Sect.~\ref{FRRT}) that function words, which belong to {\em others}, are among the most frequent words in any text. Words belonging to {\em nouns} and {\em verbs}, in contrast, are more specific to the contents developed in the text, and are mostly relegated to the low-frequency range. 

Along this same line of argument, we can outline an explanation for the fact that the rank mean anomalies for the three tags approach values closer to zero as the whole vocabulary increases in size, as shown in Fig.~\ref{fig4}. In fact, as a literary work becomes longer and its vocabulary grows \cite{nos}, the fraction of functional words in the tag {\em others} should decrease. In sufficiently large vocabularies, words such as adjectives and adverbs, which also belong to {\em others} but are grammatically linked to the other two classes, are expected to overcome in number the relatively limited set of function words. Indeed, the number of function words in English has been estimated to be around 500 \cite{Caplan}.

To get a similar estimate for the log-frequency mean anomaly $\Delta_{g'}$, besides choosing a sub-vocabulary which is uniformly distributed between ranks $r_{\min}$ and $r_{\max}$, it is necessary to advance a hypothesis for the frequency-rank relationship in the whole vocabulary from which the sub-vocabulary is selected. Quite naturally, we propose a Zipf-like relation of the form $f_r =  f_1 r^{-z}$, with a generic Zipf exponent $z$. The constant $f_1$ gives the number of occurrences of the most frequent word. Using Eq.~(\ref{rmod}), we get for the corresponding log-frequency 
\begin{equation} \label{gest}
    g_{r'} = \log f_1 -z \log \left[ r_{\min} + \frac{r'-1}{V'-1} \left(r_{\max}- r_{\min}\right) \right],
\end{equation}
as a function of $r'$. Since, from our proposal for $f_r$, it is not possible to give an explicit expression for the average $\langle g_{r'} \rangle$, we assume that it can be approximately evaluated as 
\begin{equation}
    \langle g_{r'} \rangle = \log f_1 -z \log \left[ 1 + \frac{r'-1}{V'-1} \left(V- 1\right) \right],
\end{equation}
namely, the same expression as in Eq.~(\ref{gest}) for a sub-vocabulary which is uniformly distributed along the entire vocabulary --i.e., with $r_{\min} =1$ and $r_{\max =V}$; 
cf.~Eq.~(\ref{rpmod}).     

Replacing these forms of $g_{r'}$ and $\langle g_{r'} \rangle$ in the second of Eqs.~(\ref{avanom}) and assuming, as for the rank mean anomaly, that the standard deviation $\sigma_{g'}$ can be approximated by an $r'$-independent effective value $\sigma_{g'}^{\rm eff}$, we get
\begin{equation} \label{Dgest}
    \Delta_{g'} = \frac{z}{\sigma_{g'}^{\rm eff}} \left( \log V - \frac{r_{\max} \log r_{\max} - r_{\min} \log r_{\min}}{r_{\max}-r_{\min}}\right).
\end{equation}
To get this result, we have approximated the sum in the second of Eqs.~(\ref{avanom}), and supposed that $V, V'\gg 1$, and that all words in the sub-vocabulary have frequencies above $f_r=10$. 

The rightmost panel in Fig.~\ref{fig7} shows the regions of positive and negative $\Delta_{g'}$, as estimated by Eq.~(\ref{Dgest}), on the plane spanned by $r_{\min}$ and $r_{\max}$. Taking into account the results presented in Fig.~\ref{fig5}, we see that the analysis of the sign of  $\Delta_{g'}$ qualitatively confirms the conclusions obtained from that of $\Delta_{r'}$. Concretely, the mostly positive values of $\Delta_{g'}$ for the tag {\em others} is a consequence of relative small $r_{\min}$ and $r_{\max}$. Words in this tag, therefore, accumulate in the range of large frequencies. In contrast, words in {\em nouns} and {\em verbs} tend to be distributed towards larger ranks and lower frequencies. 

While the same argument advanced in connection with $\Delta_{r'}$ may explain why the log-frequency mean anomaly decreases as $V$ grows --as observed, at least, for {\em others} and {\em nouns}-- an explanation for its small values for small $V$ is still pending. The overall non-monotonic dependence of $\Delta_{g'}$ on the vocabulary size for {\em other} and {\em nouns} remains an open problem. 

\section{Summary and conclusion} \label{S5}

In this contribution, we have studied the statistics of word frequencies and ranks within grammatical classes in a corpus of $75$ literary works in English. The words of each text were tagged using an automatized  analyzer of natural language, and then grouped into three classes: {\em nouns}, {\em verbs}, and {\em others}. For the sub-vocabulary formed by the words of each class, in each text we have characterized  the changes in rank and the frequency-rank relationship with respect to the whole vocabulary. To assess the statistical significance of the results, we carried out a comparison with a null hypothesis, where the sub-vocabularies were assumed to be randomly distributed in the list of words ranked by frequency of the whole text.

Overall, the quantifiers which we have defined to measure differences with the null hypothesis have revealed significant departures for the three grammatical classes. More importantly, these departures are different for each class, which indicates that the distribution of words in the whole ranking depends on their grammatical function. This points to the fact that the frequency-rank relationships within each sub-vocabulary may contain relevant linguistic information.    

In order to outline a preliminary interpretation for our results, we introduced a model sub-vocabulary formed by words evenly distributed in a defined portion of the whole list of ranked words. For this model, and within suitable approximations, we can explicitly estimate the difference with the null hypothesis, and thus compare with results for the real texts. Our conclusion is that, statistically speaking, words belonging to {\em others} have higher frequencies than in {\em nouns} and {\em verbs}. This difference, however, tends to fade out as the whole vocabularies become richer --i.e. for longer texts. A full explanation of the features observed in our quantifiers would nevertheless require expert analysis from the side of linguistics which, we acknowledge, we are not able to provide here.

Our study complements a recent contribution  where we have studied the statistics of appearance of new words along a text (Heaps' law),  discerning between the same three grammatical classes as here, and also comparing with a random-like null hypothesis \cite{nos}. There, we concluded that the appearance of  {\em verbs} and {\em others} are respectively more and less retarded with respect to the expectation by chance, while {\em nouns} approximately follow the null hypothesis. According to the present results, the ordering  {\em verbs}-{\em nouns}-{\em others} seems to also show up in the anomalies of the frequency-rank relationships. 

Much as our recent Heaps' analysis of tagged texts, the present results on frequency-rank relationship within grammatical classes add to the characterization of statistical regularities with a role in scientific and technological developments in natural language processing, understanding, and learning \cite{lu2020supervised,namazifar2020language} as well as in text-mining and information-retrieval techniques \cite{tm1,tm2}. In similar contexts, techniques inspired by Statistical Physics may contribute to the rocketing current interest on artificial intelligence \cite{Nat}.

\section*{Acknowledgements}
A. C. acknowledges financial support by grants from CONICET (PIP 112-2015-010028), FonCyT-ANPCyT (PICT 2017-0973),  MinCyT C\'ordoba (PID PGC 2018), and SeCyT–UNC\'ordoba, Argentina.

\end{document}